\documentclass[letterpaper]{article} 
\usepackage{aaai23v}  
\usepackage{times}  
\usepackage{helvet}  
\usepackage{courier}  
\usepackage[hyphens]{url}  
\usepackage{graphicx} 
\usepackage{caption} 
\usepackage{algorithm}
\usepackage{algorithmic}

\usepackage[switch]{lineno}

\usepackage{newfloat}
\usepackage{listings}
\DeclareCaptionStyle{ruled}{labelfont=normalfont,labelsep=colon,strut=off} 
\floatstyle{ruled}
\newfloat{listing}{tb}{lst}{}
\floatname{listing}{Listing}
\usepackage{hyperref}
\hypersetup{
    colorlinks=true,
    urlcolor=magenta
}
\usepackage{tikz}
\usepackage{comment}
\usepackage{amsmath,amssymb}
\usepackage{colortbl}
\usepackage{wrapfig}
\usepackage{bm}
\usepackage{nicematrix}
\usepackage{enumitem}
\usepackage{caption}
\usepackage{adjustbox}
\usepackage{makecell}
\usepackage{booktabs}
\usepackage{multirow}
\usepackage{multicol}
\usepackage{xcolor}
\usepackage{pifont}
\usepackage{subcaption} 
\newcommand{\tikzcircle}[2][red,fill=red]{\tikz[baseline=-0.5ex]\draw[#1,radius=#2] (0,0) circle ;}
\usepackage{soul}
\setstcolor{red}

\newcommand{\ieno}{\textit{i}.\textit{e}.}
\newcommand{\egno}{\textit{e}.\textit{g}.} 
\newcommand{\etcno}{\textit{etc}} 

\definecolor{ForestGreen}{RGB}{0, 179, 45}
\definecolor{myred}{RGB}{243, 45, 136}
\definecolor{Gray}{RGB}{147, 171, 247}

\newcommand{\tcg}{\textcolor{ForestGreen}}

\newcolumntype{a}{>{\columncolor{Gray}}c}
\newcolumntype{b}{>{\columncolor{white}}c}

\newcommand{\ours}{{ATMNet}}
\newcommand{\ourcore}{{ATM}}
\newcommand{\ourop}{{ATM}}

\newcommand{\oursxt}{ATMNet-xT}
\newcommand{\ourst}{ATMNet-T}

\newcommand{\ourss}{ATMNet-S}
\newcommand{\oursb}{ATMNet-B}
\newcommand{\oursl}{ATMNet-L}

\newcommand{\imnt}{ImageNet}
\newcommand{\imntk}{ImageNet-1K}

\newcommand{\rc}{blue!8}

\newcommand{\miou}{mIoU}
\newcommand{\sota}{state-of-the-art}
\colorlet{lgray}{gray!15}
\colorlet{lblue}{blue!8}

\usepackage[utf8]{inputenc}
\title{Active Token Mixer}
\author {
    Guoqiang Wei \textsuperscript{\rm 1}\thanks{Equal contribution. This work was done when Guoqiang Wei was an intern at MSRA.}\quad
    Zhizheng Zhang \textsuperscript{\rm 2\,$\ast$}\quad
    Cuiling Lan \textsuperscript{\rm 2}\quad
    Yan Lu \textsuperscript{\rm 2}\quad
    Zhibo Chen \textsuperscript{\rm 1}
}
\affiliations {
    \textsuperscript{\rm 1} University of Science and Technology of China\\
    \textsuperscript{\rm 2} Microsoft Research Asia\\
    wgq7441@mail.ustc.edu.cn\quad\{zhizzhang, culan, yanlu\}@microsoft.com\quad chenzhibo@ustc.edu.cn
}

\begin{document}

\maketitle

\begin{abstract}
The three existing dominant network families, i.e., CNNs, Transformers, and MLPs, differ from each other mainly in the ways of fusing spatial contextual information,
leaving designing more effective token-mixing mechanisms at the core of backbone architecture development. 
In this work, we propose an innovative token-mixer, dubbed Active Token Mixer (\textbf{ATM}), to actively incorporate flexible contextual information distributed across different channels from other tokens into the given query token. 
This fundamental operator actively predicts where to capture useful contexts and learns how to fuse the captured contexts with the query token at channel level. 
In this way, the spatial range of token-mixing can be 
expanded to a global scope with limited computational complexity, where the way of token-mixing is reformed. 
We take ATM as the primary operator and assemble ATMs into a cascade architecture, dubbed \textbf{ATMNet}.
Extensive experiments demonstrate that ATMNet is generally applicable and comprehensively surpasses different families of SOTA vision backbones by a clear margin on a broad range of vision tasks, including visual recognition and dense prediction tasks. Code is available at \url{https://github.com/microsoft/ActiveMLP}.
\end{abstract}

\section{Introduction}
Convolutional neural networks (CNNs) \cite{krizhevsky2012imagenetalexnet,simonyan2014veryVGG,szegedy2015GoogleNet,szegedy2016InceptionV3,szegedy2017inceptionV4,chollet2017xception,he2016deepresnet,huang2017DenseNet,xie2017ResNeXt,zhang2020resnest} serve as the most prevalent vision backbones for a long time. Inspired by the successes in Natural Language Processing (NLP), DETR \cite{carion2020DETR} and ViT \cite{alexander2021vit} introduce self-attention based model, \ieno, Transformer, into computer vision. 
Afterwards, Transformers spring up and make splendid breakthroughs 
on various vision tasks \cite{liu2021swin,he2021TransReID,wang2021TransTrack,xie2021SegFormer,cheng2021MaskFormer,lin2021PoseTransformer,he2021MAE}. 
Most recently, the multi-layer perceptrons (MLPs) based architectures \cite{tolstikhin2021mlpmixer,lian2021asmlp} have regained their light and been demonstrated capable of achieving stunning results on various vision tasks \cite{touvron2021resmlp,tolstikhin2021mlpmixer,chen2022cyclemlp,lian2021asmlp,zhang2021morphmlp,tang2021sparsesMLP}.


Those three categories of architectures differ from each other mainly in their different ways of token mixing. For different architectures, we uniformly refer to each feature vector as one token.
CNN-based architectures~\cite{simonyan2014veryVGG,he2016deepresnet,huang2017DenseNet} mix tokens locally within a sliding window of a fixed shape. Transformer-based architectures \cite{alexander2021vit,touvron2021deit,wang2021pyramidpvt} perform message passing from tokens in the global scope into the query token based on the pairwise attentions commonly modeled by the affinities between tokens in the embedding space. MLP-based architectures mostly enable spatial information interaction through the fully connected layers across all tokens \cite{tolstikhin2021mlpmixer,touvron2021resmlp,hou2021visionvip,tang2021sparsesMLP} or certain tokens selected with hand-crafted rules in a deterministic manner \cite{chen2022cyclemlp,zhang2021morphmlp,wang2022shift,yu2022s2MLP,lian2021asmlp,tang2021imagewavemlp}. However, the fully connected layer across all tokens makes the model unable to cope with the inputs of variable resolutions.
Adopting manually designed rules for token selection 
relaxes this constraint on fixed resolutions by restricting token mixing within a \textit{deterministic} region, but sacrificing the adaptability to various visual contents of diverse feature patterns.

In this work, we first revisit the token mixing mechanisms in dominant types of architectures from a unified perspective, then propose a novel Active Token Mixer (\ourop). 
As an innovative basic operator, \ourop~considers two properties of the learned features to actively select the tokens for mixing: 
1) the semantics in different spatial positions may correspond to diverse scales and deformations; 
2) different semantic attributes of a token would distribute in different channels \cite{bau2020understanding,wu2021stylespace}.
As illustrated in Fig.~\ref{fig: atm_module} (a), for a query, \ourop~actively predicts the locations offsets of tokens whose information should be incorporated for interaction.
Particularly, \ourop~predicts the respective offset \emph{channel-wisely} to select the context elements which are then recomposed to a new token. This empowers a more adaptive and flexible information interaction across tokens.
We adopt this operation along the horizontal and vertical dimensions in parallel (Fig.~\ref{fig: atm_module} (b)), 
making such predictive context localization easier to be optimized. Then we learn to adaptively fuse the two recomposed tokens and the original query to be the output

The \ourop~can serve as a primary operator for constructing backbone architectures. To showcase this, we build a series of model variants with different model scales, named \ours-xT/T/S/B/L, respectively. \ours~shows impressive effectiveness of \ourop~on a broad range of vision tasks 
as well as favorable scalability over different model scales. Besides, \ourop~can also serve as a plug-and-play enhanced replacement 
of the conventional convolution layers in FPN \cite{lin2017feature} to enhance the pyramid feature learning for dense prediction tasks (object detection and segmentation).

Our contributions can be summarized below:
\begin{itemize}[noitemsep,nolistsep,leftmargin=*]
\item We propose Active Token Mixer (\textbf{\ourcore}), a basic operator to efficiently enable content-adaptive and flexible global scope token mixing at channel level. It expands the range and reforms the way of message passing.
\item We build an efficient vision backbone \ours~with \ourop~as its primary ingredient for effective spatial information interaction. For the commonly used neck structure FPN, we build an enhanced FPN, \ieno, ATMFPN, powered by ATM, for dense prediction tasks.
\item \ours~achieves strong performance over different model scales and across various vision tasks. For image classification, only trained on ImageNet-1K, 
\ours~achieves 82.0\% top-1 accuracy with 27M parameters and reaches 84.8\% when scaling up to 76M. Moreover, \ours~outperforms recent prevalent backbones on dense prediction tasks by a significant margin with comparable or even less parameters and computation cost.
\end{itemize}

\section{Related Work}

\subsection{CNN based Models}

Convolutional neural networks (CNNs) have been the mainstream architectures in computer vision for a long time.
The CNN model is originally presented in \cite{lecun1998gradientlenet} for document recognition. Beginning with the significant success of AlexNet \cite{krizhevsky2012imagenetalexnet} in ILSVRC 2012, various CNN-based architectures are designed or searched, \egno, Inception \cite{szegedy2015GoogleNet,szegedy2016InceptionV3,szegedy2017inceptionV4}, VGG \cite{simonyan2014veryVGG}, ResNet \cite{he2016deepresnet}, DenseNet \cite{huang2017DenseNet}, ResNeXt \cite{xie2017ResNeXt}, EfficientNet \cite{tan2019EfficientNet}, MNASNet, \cite{tan2019MNASNet} and others \cite{wang2020HRNet,ding2021RepVGG,liu2022convnext}. In addition, there are a series of works dedicated to improving the convolution layers from different perspectives, \egno, depthwise separable convolution \cite{chollet2017xception,howard2017Mobilenetv1,sandler2018Mobilenetv2} for reduced computation costs and deformable convolution \cite{dai2017deformabledcn,zhu2019DCNV2} for objects of diverse shapes.
It is noteworthy that the deformable convolution also allows learnable token selection for token mixing but ignores the semantic differences across channels \cite{bau2020understanding,wu2021stylespace} and
usually suffers from optimization difficulties \cite{chan2021understanding}.

\subsection{Self-attention based Models}

\cite{alexander2021vit} firstly introduces a pure self-attention based backbone to computer vision, \ieno, ViT, which achieves promising performance on image classification especially trained with extremely large-scale data. 
\cite{touvron2021deit} improves the training strategy of ViT and proposes a knowledge distillation method, which helps ViT achieve higher performance 
trained only on \imnt. 
Afterwards, various works endeavor to explore efficient vision Transformer architectures,
\egno, PVT \cite{wang2021pyramidpvt,wang2021pvtv2}, Swin \cite{liu2021swin,liu2021swinv2}, Twins \cite{chu2021twins}, MViT \cite{yan2022MViT,li2021MViT2}, and others \cite{chu2021CPVT,dong2021cswin,ali2021XCiT,touvron2021CaiT,yang2021Focal,bertasius2021TimeSformer,li2022UniFormer}. 
Transformer also presents its superiority on various tasks, \egno, object detection \cite{carion2020DETR,zhu2021deformabledetr}, segmentation \cite{cheng2021MaskFormer,cheng2021Mask2Former,xie2021SegFormer,zheng2021SETR}, pose estimation \cite{lin2021PoseTransformer}, tracking \cite{wang2021TransTrack,chen2021TransT} and GAN \cite{jiang2021TransGAN,xu2021STransGAN}. 

\subsection{MLP-like Models}

Recently, MLP-like models have been reinvigorated. The pioneering works MLP-Mixer \cite{tolstikhin2021mlpmixer} and ResMLP \cite{touvron2021resmlp} stack two types of MLP layers, \ieno, token-mixing MLP and channel-mixing MLP, alternately. The token-mixing MLP enables spatial information interaction over all tokens while the channel-mixing MLP mixes information across all channels within each token.
ViP \cite{hou2021visionvip} and sMLP \cite{tang2021sparsesMLP} encode the feature representations along two axial dimensions to improve MLPs' efficiency and capability. Shift \cite{wang2022shift}, ASMLP \cite{lian2021asmlp} and S$^2$MLP \cite{yu2022s2MLP} perform spatial information mixing with spatial shift operations along different dimensions. CycleMLP \cite{chen2022cyclemlp}, WaveMLP \cite{tang2021imagewavemlp} and MorphMLP \cite{zhang2021morphmlp} restrict the spatial information interaction within \textit{hand-craft fixed local} windows in a \textit{deterministic} way. As opposed to them, our \ourop~achieves \textit{a learnable content-adaptive token-mixing}, which considers the diverse semantics attributed in different channels and spatial positions with global receptive fields, so that it can attain high flexibility and strong modeling capacity.

\begin{figure*}[t]
	\centering
	\includegraphics[width=0.99\textwidth]{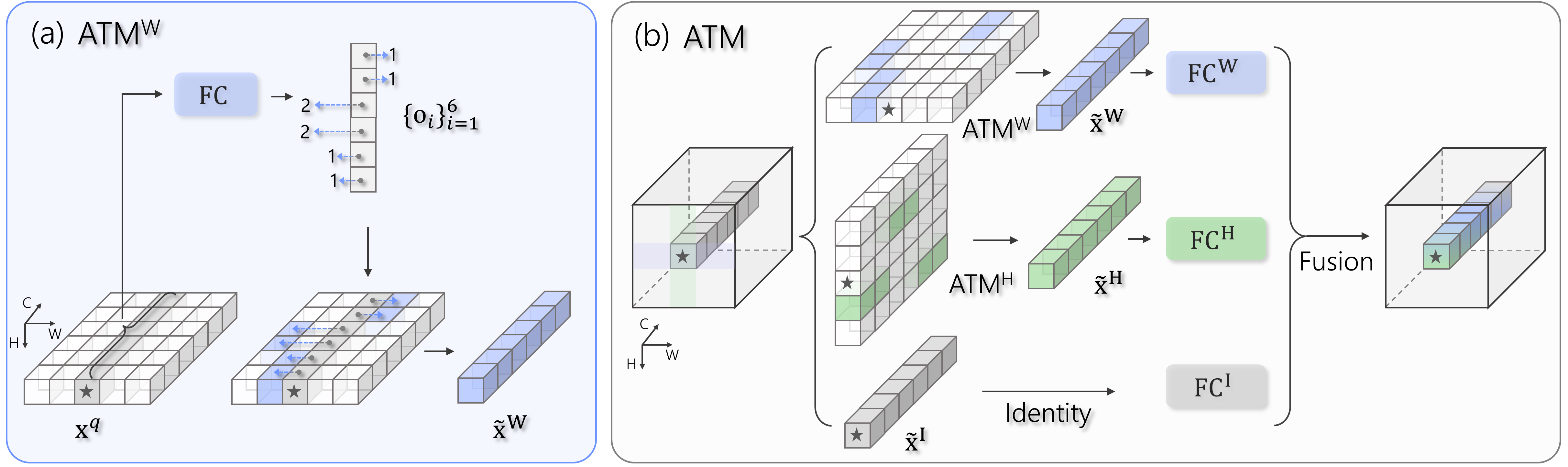}
    \caption{Illustration of our proposed Active Token Mixer (ATM). 
    (a) ATM along the horizontal (width) dimension.
    For a query $\mathbf{x}^q$, ATM actively captures the useful contexts by recomposing the elements from selected tokens into $\mathbf{\tilde{x}}^{W}\in \mathbb{R}^C$ based on the learned channel-wise offsets.
    (b) ATM module consisting 
    of ATM$^W$ along horizontal dimension, ATM$^H$ along vertical dimension, and the identity branch ATM$^I$. The two recomposed tokens ($\mathbf{\tilde{x}}^{W}$, $\mathbf{\tilde{x}}^{H}$) and the original  $\mathbf{\tilde{x}}^{I}$ are then adaptively fused after being embedded by $FC^{\{W, H, I\}}$. 
    }	
    \label{fig: atm_module}
\end{figure*}
\section{Method}

\subsection{A Unified Perspective of Token Mixing}
\label{sec: UnifiedPerspective}

For most prevailing model architectures, the input image is first patchified into a feature tensor $\mathbf{X}\in \mathbb{R}^{H\times W \times C}$ with the height $H$, the width $W$ and the number of channels $C$.
In vision tasks, token mixing is especially critical since the contextual information is inevitably required for understanding visual semantics.
Before introducing our proposed method, we firstly review different token mixing mechanisms in the literature from a unified perspective.
Mathematically, we formulate token mixing with a unified function:
\begin{align}
    f(\mathbf{X})\vert_{\mathbf{x}^q} = \sum_{k\in\mathcal{N}(\mathbf{x}^q)} 
    \bm{\omega}^{k\rightarrow q} \ast
    g(\mathbf{x}^k),
    \label{eq: general_formulation}
\end{align}
where $\mathbf{x}^q$ denotes the query token while $\mathcal{N}(\mathbf{x}^q)$ refers to a set of its contextual tokens.
$\bm{\omega}^{k\rightarrow q}$ is the weight determining the degree of message passing from $\mathbf{x}^k$ to $\mathbf{x}^q$.
$g(\cdot)$ is an embedding function. 
$\ast$ is a unified representation for element-wise or matrix multiplication.

For conventional CNNs, $g(\cdot)$ is an identity function, and $\bm{\omega}^{k\rightarrow q}\in \mathbb{R}^{C\times C}$ corresponds to the convolutional kernels shared for different queries, and the message passing is restricted within a fixed-size sliding window $\mathcal{N}(\cdot)$. 
Transformers achieve a non-local $\mathcal{N}(\cdot)$ and adopt a computationally expensive
$\bm{\omega}^{k\rightarrow q}\in \mathbb{R}^C$ through calculating the affinity between $\mathbf{x}^k$ and $\mathbf{x}^q$ in the embedding space. 
In recent MLP-like backbones \cite{chen2022cyclemlp,touvron2021resmlp,tolstikhin2021mlpmixer,zhang2021morphmlp,lian2021asmlp,tang2021sparsesMLP,tang2021imagewavemlp}, $\mathcal{N}(\cdot)$ and $\bm{\omega}^{k\rightarrow q}$ are manually designed to perform token mixing in a \textit{deterministic} way, 
leading to the \textit{lack of content adaptivity}. In Transformers or MLPs, $g(\cdot)$ is commonly a learnable embedding function.


\subsection{Active Token Mixer}





Based on the token mixing methods detailed in Sec. \ref{sec: UnifiedPerspective}, we have two key observations:
1) For the spatial dimension, visual objects/stuffs present diverse shapes and deformations. Therefore, information mixing within a fixed-range
$\mathcal{N}(\cdot)$ \cite{touvron2021resmlp,chen2022cyclemlp,tolstikhin2021mlpmixer,lian2021asmlp} is inefficient and inadequate. 
The adaptive $\bm{\omega}^{k\rightarrow q}$ and $\mathcal{N}(\cdot)$ for message passing are desirable for extracting visual representations.
2) For the channel dimension, multiple semantic attributes carried in one token would distribute in its different channels \cite{bau2020understanding,wu2021stylespace}. The token-level message passing with  $\omega^{k\rightarrow q}\in \mathbb{R}$ shared over all channels can not treat different semantics adaptively 
and limits their full use, thus is less effective \cite{touvron2021resmlp,tolstikhin2021mlpmixer}. In this work, \textit{we pinpoint the importance of more fine-grained message passing for treating different semantics adaptively}.

\begin{table*}[t]
\footnotesize
\begin{center}

\begin{tabular}{@{}c@{}}
\begin{minipage}[b]{0.48\textwidth}
\resizebox{\textwidth}{!}{
\begin{NiceTabular}{@{}lcccc@{}}[code-before=
\rectanglecolor{\rc}{5-1}{5-5}
\rectanglecolor{\rc}{20-1}{20-5}
\rectanglecolor{\rc}{26-1}{26-5}]
\toprule
Model & Size & \makecell{\#P.(M)} & \makecell{FLOPs(G)} & \makecell{Top-1(\%)} \\

\midrule
ResNet18 \cite{he2016deepresnet}   
    & 224$^2$   & 12    & 1.8   & 69.8 \\
ResMLP-S12 \cite{touvron2021resmlp}  
    & 224$^2$   & 15    & 3.0   & 76.6 \\
CycleMLP-B1 \cite{chen2022cyclemlp} 
    & 224$^2$   & 15    & 2.1   & 78.9 \\
\oursxt     & 224$^2$   & 15    & 2.2   & \textbf{79.7} \\

\midrule
ResNet50 \cite{he2016deepresnet, wightman2021resnettimm}    
    & 224$^2$   & 26    & 4.1   & 79.8 \\
Deit-S \cite{touvron2021deit}
    & 224$^2$   & 22    & 4.6   & 79.8 \\
Deit III-S \cite{touvron2022deit3}
    & 224$^2$   & 22    & 4.6   & 81.4 \\
PVT-S \cite{wang2021pyramidpvt}
    & 224$^2$   & 25    & 3.8   & 79.8 \\
Swin-T \cite{liu2021swin} 
    & 224$^2$   & 29    & 4.6   & 81.2 \\
TwinsP-S \cite{chu2021twins}
    & 224$^2$   & 24    & 3.8   & 81.2 \\
Twins-S \cite{chu2021twins} 
    & 224$^2$   & 24    & 2.9   & 81.7 \\
ResMLP-S24 \cite{touvron2021resmlp}
    & 224$^2$   & 30    & 6.0   & 79.4 \\
ASMLP-T \cite{lian2021asmlp}
    & 224$^2$   & 28    & 4.4   & 81.3 \\
ViP-S \cite{hou2021visionvip}
    & 224$^2$   & 25    & 6.9   & 81.5 \\ 
MorphMLP-T \cite{zhang2021morphmlp}
    & 224$^2$   & 23    & 3.9   & 81.6 \\
CycleMLP-B2 \cite{chen2022cyclemlp}
    & 224$^2$   & 27    & 3.9   & 81.6 \\
Shift-T \cite{wang2022shift}
    & 224$^2$   & 29    & 4.5   & 81.7 \\
sMLP-T \cite{tang2021sparsesMLP}
    & 224$^2$  & 24    & 5.0   & 81.9 \\
\ourst      & 224$^2$   & 27    & 4.0   & \textbf{82.0} \\

\midrule   
PVT-M \cite{wang2021pyramidpvt}
    & 224$^2$  & 44    & 6.7   & 81.2 \\
TwinsP-B \cite{chu2021twins}
    & 224$^2$  & 44    & 6.7   & 82.7 \\
MorphMLP-S \cite{zhang2021morphmlp}
    & 224$^2$  & 38    & 7.0   & 82.6 \\
CycleMLP-B3 \cite{chen2022cyclemlp}
    & 224$^2$  & 38    & 6.9   & 82.6 \\
Deit III-M \cite{touvron2022deit3}
    & 224$^2$  & 39    & 8.0 & 83.0 \\    
\ourss      & 224$^2$  & 39    & 6.9   & \textbf{83.1} \\
\bottomrule
\end{NiceTabular}
}
\end{minipage}

\hspace{0.02\textwidth}

\begin{minipage}[b]{0.48\textwidth}
\resizebox{\textwidth}{!}{
\begin{NiceTabular}{@{}lcccc@{}}[code-before=
\rectanglecolor{\rc}{11-1}{11-5}
\rectanglecolor{\rc}{22-1}{22-5}
\rectanglecolor{\rc}{26-1}{26-5}]
\toprule
Model & Size & \makecell{\#P.(M)} & \makecell{FLOPs(G)} & \makecell{Top-1(\%)} \\
\midrule
PVT-L \cite{wang2021pyramidpvt} 
    & 224$^2$   & 61    & 9.8   & 81.7 \\
Swin-S \cite{liu2021swin}
    & 224$^2$   & 50    & 8.7   & 83.2 \\
Twins-B \cite{chu2021twins}
    & 224$^2$   & 56    & 8.6   & 83.2 \\
Mixer-B/16 \cite{tolstikhin2021mlpmixer} 
    & 224$^2$   & 59    & 12.7  & 76.4 \\
ViP-M \cite{hou2021visionvip}
    & 224$^2$   & 55    & 16.3  & 82.7 \\
Shift-S \cite{wang2022shift}
    & 224$^2$   & 50    & 8.8   & 82.8 \\
CycleMLP-B4 \cite{chen2022cyclemlp}
    & 224$^2$   & 52    & 10.1  & 83.0 \\[1pt]
ASMLP-S \cite{lian2021asmlp}
    & 224$^2$   & 50    & 8.5   & 83.1 \\[1pt]
MorphMLP-B \cite{zhang2021morphmlp}
    & 224$^2$   & 58    & 10.2  & 83.2 \\
\oursb      & 224$^2$   & 52    & 10.1  & \textbf{83.5} \\
\midrule
Deit-B \cite{touvron2021deit}
    & 224$^2$   & 86    & 17.5  & 81.8 \\
Deit III-B \cite{touvron2022deit3}
    & 224$^2$   & 87    & 17.5  & \textbf{83.8} \\
Swin-B \cite{liu2021swin}
    & 224$^2$   & 88    & 15.4  & 83.5 \\[1pt]
S$^2$MLP-W \cite{yu2022s2MLP}
    & 224$^2$   & 71    & 14.0  & 80.0 \\
CycleMLP-B5 \cite{chen2022cyclemlp}
    & 224$^2$   & 76    & 15.3  & 83.1 \\
ViP-L \cite{hou2021visionvip}
    & 224$^2$   & 88    & 24.4  & 83.2 \\
Shift-B \cite{wang2022shift}
    & 224$^2$   & 89    & 15.6  & 83.3 \\[1pt]
ASMLP-B \cite{lian2021asmlp}
    & 224$^2$   & 88    & 15.2  & 83.3 \\
MorphMLP-L \cite{zhang2021morphmlp}
    & 224$^2$   & 76    & 12.5  & 83.4 \\
sMLP-B \cite{tang2021sparsesMLP} 
    & 224$^2$   & 66    & 14.0  & 83.4 \\
\oursl      & 224$^2$   & 76    & 12.3  & \textbf{83.8} \\
\hline
ViT-B/16$\uparrow$ \cite{alexander2021vit}
            & 384$^2$   & 86    & 55.4  & 77.9 \\
Deit-B$\uparrow$ \cite{touvron2021deit}
            & 384$^2$   & 86    & 55.4  & 83.1 \\
Swin-B$\uparrow$ \cite{liu2021swin}
            & 384$^2$   & 88    & 47.1  & 84.5 \\
\oursl$\uparrow$ 
            & 384$^2$   & 76    & 36.4  & \textbf{84.8} \\
\bottomrule
\end{NiceTabular}
}
\end{minipage}
\end{tabular}
\captionof{table}{Comparisons with \sota~models on ImageNet-1K without extra data. All models are trained with input size of 224$\times$224, except $\uparrow$ with 384$\times$384.}
\label{table: int_sota}
\end{center}
\end{table*}

To address the aforementioned limitations in existing token-mixing methods, we propose Active Token Mixer (\ourop) as shown in Fig. \ref{fig: atm_module}. 
It first predicts the relative locations of useful contextual tokens along each direction at channel level, then learns to fuse the contextual tokens and query token.
These two steps correspond to learn \textbf{where} the useful context tokens locate in and \textbf{how} to fuse them with the original information, respectively.

Drawing on the success of multi-branch design in \cite{hou2021visionvip,chen2022cyclemlp,lian2021asmlp}, we propose a three-branch architecture for facilitating the context localization along different directions.
Two branches are responsible for recomposing tokens into a new one along two axial directions separately as shown in Fig. \ref{fig: atm_module} (b). In addition, we adopt an identity branch to preserve the original query information.
The two recomposed tokens and query are further mixed as the final output. 

\subsubsection{ATM along the horizontal dimension}
We illustrate the \ourop~along the horizontal (width) dimension, denoted by ATM$^W$, in Fig. \ref{fig: atm_module} (a).
Given the query $\mathbf{x}^q \in \mathbb{R}^{C}$ (marked with $\star$), we first feed it into a FC layer to adaptively predict $C$ offsets $\mathcal{O}=\{ o_i \}_{i=1}^C$ for context localization.  
Note that we impose no constraint on the offset generation, thus $\mathcal{N}(\mathbf{x}^q)$ is allowed to be extended to all spatial positions along the horizontal direction. In this way, \ourop~can incorporate the information from the global scope, wherever needed, into $\mathbf{x}^q$ in a flexible and active manner. The predicted offsets determine the
tokens in $\mathcal{N}(\cdot)$ per channel, which
are used to recompose the selected tokens into a new token $\mathbf{\tilde{x}}^{W}\in \mathbb{R}^C$
as output of ATM$^W$:
\begin{align}
    \mathbf{\tilde{x}}^{W} = \big[  \mathbf{X}_{\left[ i, j + o_1, 1 \right]}, \mathbf{X}_{\left[ i, j + o_2, 2 \right]}, \dots, \mathbf{X}_{\left[ i, j + o_C, C \right]} \big]^T,
\end{align}
where $\mathbf{X}_{[i,j+o,c]}$ denotes the $c^{th}$ channel element of the token at spatial position $[i, j+o]$ where $[i,j]$ is the position of $\mathbf{x}^q$. 
ATM$^W$ is capable of mixing information horizontally and globally into $\mathbf{\tilde{x}}^{W}$.

\subsubsection{ATM along the vertical dimension}
Likewise, another ATM$^H$ branch 
is adopted to recompose a token \textbf{$\mathbf{\tilde{x}}^{H}$} along the vertical (height) dimension.

\subsubsection{Fusion}
Here, we introduce how to fuse the recomposed $\mathbf{\tilde{x}}^{W}$, $\mathbf{\tilde{x}}^{H}$ and the original $\mathbf{\tilde{x}}^{I}$ 
into the final token-mixing result. 
First, we adopt three FC layers $FC^{\{W, H, I\}}$ to embed $\mathbf{\tilde{x}}^{\{W, H, I\}}$
to $\mathbf{\hat{x}}^{\{W, H, I\}}$, respectively, which are then mixed with learned weights, formulated as:
\begin{gather}
    \hat{\mathbf{x}} = 
\bm{\alpha}^{\scriptscriptstyle W}\odot\hat{\mathbf{x}}^{\scriptscriptstyle W} +
\bm{\alpha}^{\scriptscriptstyle H}\odot\hat{\mathbf{x}}^{\scriptscriptstyle H} +
\bm{\alpha}^{\scriptscriptstyle I}\odot\hat{\mathbf{x}}^{\scriptscriptstyle I},
\end{gather}
where $\odot$ denotes element-wise multiplication. $\bm{\alpha}^{\scriptscriptstyle \left\{ W,H,I \right\}} \in \mathbb{R}^{\scriptscriptstyle C}$ are learned from the summation $\hat{\mathbf{x}}^{\scriptscriptstyle \sum}$ of $\mathbf{\hat{x}}^{\{W, H, I\}}$ 
with 
$W^{\scriptscriptstyle \left\{ W,H,I \right\}} \in \mathbb{R}^{\scriptscriptstyle C\times C}$: 
\begin{gather}
\left[ \bm{\alpha}^{\scriptscriptstyle W}, 
\bm{\alpha}^{\scriptscriptstyle H}, 
\bm{\alpha}^{\scriptscriptstyle I} \right] = 
\sigma(\left[ 
W^{\scriptscriptstyle W} \cdot \hat{\mathbf{x}}^{\scriptscriptstyle \sum},
W^{\scriptscriptstyle H} \cdot \hat{\mathbf{x}}^{\scriptscriptstyle \sum},
W^{\scriptscriptstyle I} \cdot \hat{\mathbf{x}}^{\scriptscriptstyle \sum}
\right] ),
\end{gather}
where $\sigma(\cdot)$ is a \textit{softmax} function for normalizing each channel separately. 


\subsubsection{Discussion}
Our \ourcore~has three hallmarks: 
1) \textit{Content adaptivity.} The context selection/localization 
is adaptively learned for the query token in an active way, instead of being passively determined by manual designed rules \cite{chen2022cyclemlp,lian2021asmlp,yu2022s2MLP,zhang2021morphmlp}.
2) \textit{Flexibility.} In general, different channels are characterized with different semantics. Our proposed \ourop~enables to dynamically select context tokens at the channel level from a global range $\mathcal{N}(\cdot)$, adaptive to visual contents with various scales and deformations.
3) \textit{Efficiency.} 
By incorporating contexts from $C$ tokens into the two recomposed tokens, 
the computation complexity of ATM is $\mathcal{O}\left(HWC^2\right)$, which is linear with the input resolution and is agnostic to the receptive fields, making it computation-friendly to larger-size images used in object detection and segmentation tasks.

Compared with the conventional convolutions, 
\ourcore~is able to enlarge its receptive field to global-scope flexibly with constant computation cost. Compared with the multi-head self-attention in Transformers, \ourcore~globally mixes token information per channel with the actively learned offsets, avoiding the computation-consuming attention calculation. 
\ourcore~may be reminiscent of the deformable convolution \cite{dai2017deformabledcn,zhu2019DCNV2}. 
In fact, there are two crucial differences: 1) The learned offsets in deformable convolutions are shared over all channels, without consideration on semantic differences across channels.
Our ATM can incorporate contextual information in channel wise, achieving a more flexible and fine-grained context exploitation mechanism in token mixing.
2) We decouple the learning of context localization along different directions, 
making \ours~easier to be optimized.

\subsection{Model Architectures}
\noindent
\subsubsection{ATM Block}
We build our \ours~by stacking multiple \ourop~blocks in sequence. 
Here, we introduce the architecture of an \ourop~block. 
For the output $\mathbf{X}^{\textit{l-1}}$ of the $(l-1)$-th block \ourop$^{\textit{l-1}}$, we feed it to the $l$-th block \ourop$^{l}$ for token mixing. 
Further, we use an MLP module to further modulate the feature along its channel dimension. Skip connections are adopted to facilitate the training. The entire process can be formulated as:
\begin{align}
    &\hat{\mathbf{X}}^\textit{l} = ATM^l( LN( \mathbf{X}^{\textit{l-1}} ) ) + \mathbf{X}^{\textit{l-1}}, \\
    &\mathbf{X}^l = MLP^l( LN( \hat{\mathbf{X}}^{\textit{l}} ) ) + \hat{\mathbf{X}}^{\textit{l}},
\end{align}
where $LN$ is LayerNorm \cite{ba2016LayerNorm}.

\noindent
\subsubsection{\ours} 
Following the typical hierarchical architecture designs \cite{he2016deepresnet,liu2021swin}, we provide five four-stage backbone architecture variants with different channel dimensions and numbers of the \ourcore~blocks, which are \oursxt/T/S/B/L, respectively. 
Note that the offset generation layer is shared across the tokens within each ATM branch. 
Here, the awareness of the position of query token can facilitate offsets prediction. We thus introduce one positional encoding generator (PEG) \cite{chu2021CPVT} for each stage before ATM, which helps a little for dense prediction tasks. More details are placed in the supplementary.



\noindent
\subsubsection{ATMFPN}
In addition to the strong capability of constructing vision backbones, ATM is also an enhanced alternative for conventional convolutions in convolution-based decoders for downstream tasks. We replace the convolutions in the prevailing FPN \cite{lin2017FPN}, which is widely applied as the neck for object detection and segmentation, with our \ourcore~and name this new neck as \textbf{ATMFPN}. We demonstrate the effectiveness of our {ATMFPN} in Table~\ref{table: atm_fpn}.   

\begin{table}[t]
\begin{center}
\small
\setlength{\tabcolsep}{11pt}
\begin{NiceTabular}{@{}lccc@{}}[code-before=
\rectanglecolor{\rc}{4-1}{4-15}
\rectanglecolor{\rc}{10-1}{10-15}
\rectanglecolor{\rc}{15-1}{15-15}
\rectanglecolor{\rc}{20-1}{21-15}
]
\toprule
Backbone & \#P.(M) & FLOPs(G) & mIoU(ss)  \\

\midrule
CycleMLP-B1 \cite{chen2022cyclemlp}
    & 18.9 & 32.8 & 40.8 \\
WaveMLP-T \cite{tang2021imagewavemlp}
    & 19.3 & - & 41.2 \\
\oursxt
    & 19.1 & 33.1 & \textbf{43.0} \\
\midrule

Swin-T \cite{liu2021swin}  
    & 31.9 & 46 & 41.5 \\
Twins-S \cite{chu2021twins}
    & 28.3 & 37 & 43.2 \\
MorphMLP-T \cite{zhang2021morphmlp}
    & 26.4 & - & 43.0 \\
CycleMLP-B2 \cite{chen2022cyclemlp}
    & 30.6 & 42.0 & 43.4 \\ 
Wave-MLP-S \cite{tang2021imagewavemlp}  
    & 31.2 & - & 44.4 \\
\ourst
    & 30.9 & 42.4 & \textbf{45.8} \\
\midrule

Swin-S \cite{liu2021swin} 
    & 53.2 & 70 & 45.2 \\
Twins-B \cite{chu2021twins}
    & 60.4 & 67 & 45.3 \\
CycleMLP-B3 \cite{chen2022cyclemlp}
    & 42.1 & 57.5 & 44.3 \\
WaveMLP-M \cite{tang2021imagewavemlp}
    & 43.3 & - & 46.8 \\
\ourss 
    & 42.4 & 57.8 & \textbf{47.3} \\
\midrule

Swin-B \cite{liu2021swin}
    & 91.2 & 107 & 46.0 \\ 
Twins-L \cite{chu2021twins}
    & 103.7 & 102 & 46.7 \\
MorphMLP-B \cite{zhang2021morphmlp}
    & 59.3 & - & 45.9 \\
CycleMLP-B5 \cite{chen2022cyclemlp}
    & 79.4 & 86.0 & 45.5 \\
\oursb
    & 55.9 & 74.7 & 47.7 \\
\arrayrulecolor{white}\hline
\oursl
    & 79.8 & 86.6 &\textbf{48.1} \\
\arrayrulecolor{black}\bottomrule
\end{NiceTabular}
\captionof{table}{
Semantic segmentation results on ADE20K \texttt{val} with Semantic FPN \cite{kirillov2019semanticFPN}. FLOPs are evaluated on 512$\times$512 resolution. All backbones are pretrained on ImageNet-1K. 
}
\label{table: ade_fpn}
\end{center}
\end{table}

\begin{table}[t]
\begin{center}
\sethlcolor{lgray}
\small
\resizebox{0.475\textwidth}{!}{
\begin{NiceTabular}{@{}lcccc@{}}[code-before=
\rectanglecolor{\rc}{2-1}{2-15}
\rectanglecolor{\rc}{8-1}{8-15}
\rectanglecolor{\rc}{14-1}{14-15}
\rectanglecolor{\rc}{20-1}{21-15}
]
\toprule
Backbone & \#P.(M) & FLOPs(G) & mIoU(ss) & \miou(ms) \\
\midrule

\oursxt
    & 45 & 889 & 44.3 & 45.4 \\ 
\midrule

Swin-T \cite{liu2021swin}
    & 60 & 945 & 44.5 & 45.8 \\ 
Twins-S \cite{chu2021twins}
    & 54 & 931  & 46.2 & 47.1 \\
ConvNeXt-T \cite{liu2022convnext}
    & 60 & 939 & - & 46.7 \\
ASMLP-T \cite{lian2021asmlp}
    & 60 & 937 & - & 46.5 \\ 
CycleMLP-T \cite{chen2022cyclemlp}
    & 60 & 937 & - & 47.1 \\ 
\ourst
    & 57 & 927 & \textbf{46.5} & \textbf{47.6} \\ 
\bottomrule

Swin-S \cite{liu2021swin}
    & 81 & 1038 & 47.6 & 49.5 \\ 
Twins-B \cite{chu2021twins}
    & 89 & 1078 & 47.7 & 48.9 \\
ConvNeXt-S \cite{liu2022convnext}
    & 82 & 1027 & - & 49.6 \\
ASMLP-S \cite{lian2021asmlp}
    & 81 & 1024 & - & 49.2 \\ 
CycleMLP-S \cite{chen2022cyclemlp}
    & 81 & 1024 & - & 49.6 \\ 
\oursb
    & 82 & 1055 & \textbf{48.7} & \textbf{49.8} \\
\arrayrulecolor{black}\midrule

Swin-B \cite{liu2021swin}
    & 121 & 1188 & 48.1 & 49.7 \\ 
Twins-L \cite{chu2021twins}
    & 133 & 1236 & 48.8 & 50.2 \\
ConvNeXt-B \cite{liu2022convnext}
    & 122 & 1170 & - & 49.9 \\
ASMLP-B \cite{lian2021asmlp}
    & 121 & 1166 & - & 49.5 \\ 
CycleMLP-B \cite{chen2022cyclemlp}
    & 121 & 1166 & - & 49.7 \\ 
\ourss
    & 69 & 988 & 48.4 & 49.5 \\
\arrayrulecolor{white}\hline
\oursl
    & 108 & 1106 & \textbf{50.1} & \textbf{51.1} \\
\arrayrulecolor{black}\bottomrule
\end{NiceTabular}
}
\captionof{table}{Semantic segmentation results on ADE20K \texttt{val} with UperNet \cite{xiao2018unifiedupernet}. FLOPs are evaluated on 512$\times$2048 resolution. All backbones are pretrained on \imntk.
}
\label{table: ade_upernet}
\end{center}
\end{table}

\section{Experiments}


\subsection{ImageNet-1K Classification}
\label{sec: int_cls}

\noindent
\subsubsection{Settings}
We train our models on the \imntk~dataset \cite{deng2009imagenet} from scratch, which contains 1.2M training images and 50K validation images evenly spreading 1,000 categories. We report the top-1 accuracy on the validation set following the standard practice in this community \cite{liu2021swin,liu2022convnext,wang2021pyramidpvt,chen2022cyclemlp}. Our implementation is established with \texttt{PyTorch} \cite{pytorch} and base on the \texttt{timm} \cite{rw2019timm} repository. For fair comparisons, our training strategy is mostly inherited from DeiT \cite{touvron2021deit}, which includes RandAugment \cite{cubuk2020randaugment}, Mixup \cite{zhang2017mixup}, Cutmix \cite{yun2019cutmix}, Random erasing \cite{zhong2020randomerasing} and stochastic depth \cite{huang2016deepdpr}. The optimizer is AdamW \cite{loshchilov2017decoupledadamw} with the momentum of 0.9 and weight decay of $5\times10^{-2}$ by default. All models are trained with input size of 224$\times$224 for 300 epochs with 5-epoch warm-up and batch size of 1024. For 384$\times$384 resolution, we finetune the models for 30 epochs with the learning rate of $5e-6$. More details are shown in Table \ref{table: supp_int_setting}. 

\begin{table*}[h]
\begin{center}
\small
\begin{NiceTabular}{@{}lcccccccccccccc@{}}[code-before=
\rectanglecolor{\rc}{7-1}{8-15}
\rectanglecolor{\rc}{13-1}{15-15}
]
\toprule
\multirow{2}{*}{Backbone} & \Block{2-1}{\makecell{\#Params.\\(M)}} & \multirow{2}{*}{\makecell{FLOPs\\{(G)}}} & \multicolumn{6}{c}{\normalsize RetinaNet 1$\times$} & \multicolumn{6}{c}{\normalsize RetinaNet 3$\times$ MS} \\
& & 
& \textbf{AP$^b$} & AP$^b_{50}$ & AP$^b_{75}$ 
& AP$_S$ & AP$_{M}$ & AP$_{L}$
& \textbf{AP$^b$} & AP$^b_{50}$ & AP$^b_{75}$ 
& AP$_S$ & AP$_{M}$ & AP$_{L}$ \\
\midrule

Swin-T \cite{liu2021swin} & 39 & 245 
    & 41.5 & 62,1 & 44.2 & 25.1 & 44.9 & 55.5
    & 45.0 & 65.9 & 48.4 & 29.7 & 48.9 & 58.1\\ 
Twins-S \cite{chu2021twins} & 34 & 286
    & 43.0 & 64.1 & 46.0 & \textbf{27.5} & 46.3 & 57.3 
    & 45.2 & 66.5 & 48.6 & 30.0 & 48.8 & 58.9 \\
CycleMLP-B2 \cite{chen2022cyclemlp} & 37 & 231
    & 40.9 & 61.8 & 43.4 & 23.4 & 44.7 & 53.4
    & - & - & - & - & - & - \\ 
WaveMLP-S \cite{tang2021imagewavemlp} & 37 & 231
    & 43.4 & 64.4 & 46.5 & 26.6 & 47.1 & 57.1
    & - & - & - & - & - & -  \\ 
\oursxt & 25 & 196
    & 41.2 & 62.3 & 43.7 & 24.4 & 45.3 & 54.2 
    & 43.7 & 64.7 & 46.7 & 28.9 & 47.4 & 57.4 \\ 
\arrayrulecolor{white}\hline\arrayrulecolor{black}
\ourst & 37 & 233
    & \textbf{43.6} & \textbf{64.9} & \textbf{46.8} & 27.2 & \textbf{47.5} & \textbf{57.9}
    & \textbf{45.8} & \textbf{66.9} & \textbf{49.2} & \textbf{29.9} & \textbf{49.3} & \textbf{59.9 }\\ 
\midrule

Twins-L \cite{chu2021twins} & 111 & 528
    & 45.7 & 67.1 & 49.2 & - & - & - 
    & - &- &- &- &- &- \\
Swin-B \cite{liu2021swin} & 98 & 477
    & 44.7 & 65.9 & 47.8 & - & - & - 
    & 45.8 & 66.4 & 49.1 & 29.9 & 49.4 & 60.3 \\
CycleMLP-B5 \cite{chen2022cyclemlp}  & 86 & 402
    & 42.7 & 63.3 & 45.3 & 24.1 & 46.3 & 57.4  
    & - &- &- &- &- &- \\
WaveMLP-B \cite{tang2021imagewavemlp} & 66 & 334
    & 44.2 & 65.1 & 47.1 & 27.1 & 47.8 & 58.9 
    & - &- &- &- &- &- \\
\ourss & 48 & 293
    & 45.4 & 66.7 & 48.5 & 28.3 & 49.7 & 59.7 
    & {46.3} & 68.0 & 49.5 & 30.4 & 50.6 & 59.9 \\
\arrayrulecolor{white}\hline\arrayrulecolor{black}
\oursb & 62 & 359
    & 45.6 & 67.2 & 48.9 & 28.9 & 49.6 & 60.5 
    & {47.7} & 69.2 & 50.9 & \textbf{33.1} & 51.6 & 61.8 \\
\arrayrulecolor{white}\hline\arrayrulecolor{black}
\oursl & 86 & 405
    & \textbf{46.1} & \textbf{67.4} & \textbf{49.4} & \textbf{29.9} & \textbf{50.1} & \textbf{61.0} 
    & \textbf{48.1} & \textbf{69.5} & \textbf{51.8} & 31.7 & \textbf{52.1} & \textbf{63.0} \\

\bottomrule 
\end{NiceTabular}
\captionof{table}{Object detection results on COCO \texttt{val2017} with RetinaNet \cite{lin2017FocallossRetinaNet} 1$\times$ and 3$\times$ MS. FLOPS are evaluated with resolution 800$\times$1280.
}
\label{table: coco_retina_all}
\end{center}
\end{table*}
\begin{table*}[!h]
\begin{center}
\small
\begin{NiceTabular}{@{}lcccccccccccccc@{}}[code-before=
\rectanglecolor{\rc}{9-1}{10-15}
\rectanglecolor{\rc}{16-1}{18-15}
]
\toprule
\multirow{2}{*}{Backbone} & \Block{2-1}{\makecell{\#Params.\\(M)}} & \multirow{2}{*}{\makecell{FLOPs\\{(G)}}} & \multicolumn{6}{c}{\normalsize Mask R-CNN 1$\times$} & \multicolumn{6}{c}{\normalsize Mask R-CNN 3$\times$ MS} \\
& & & \textbf{AP$^b$} & AP$^b_{50}$ & AP$^b_{75}$ & \textbf{AP$^m$} & AP$^m_{50}$ & AP$^m_{75}$ & \textbf{AP$^b$} & AP$^b_{50}$ & AP$^b_{75}$ & \textbf{AP$^m$} & AP$^m_{50}$ & AP$^m_{75}$ \\

\arrayrulecolor{black}\midrule\arrayrulecolor{black}

Swin-T \cite{liu2021swin} & 48 & 264 
    & 42.2 & 64.6 & 46.2 & 39.1 & 61.6 & 42.0 
    & 46.0 & 68.2 & 50.2 & 41.6 & 65.1 & 44.8 \\ 
ConvNeXt-T \cite{liu2022convnext} & 48 & 262 
    &- &- &- &- &- &-
    & 46.2 & 67.9 & 50.8 & 41.7 & 65.0 & 44.9 \\
Twins-S \cite{chu2021twins} & 44 & 228 
    & 43.4 & 66.0 & 47.3 & 40.3 & 63.2 & 43.4 
    & 46.8 & \textbf{69.2} & 51.2 & 42.6 & 66.3 & 45.8 \\ 
ASMLP-T \cite{lian2021asmlp} & 48 & 260
    &- &- &- &- &- &-
    & 46.0 & 67.5 & 50.7 & 41.5 & 64.6 & 44.5 \\ 
CycleMLP-B2 \cite{chen2022cyclemlp} & 47 & 250  
    & 42.1 & 64.0 & 45.7 & 38.9 & 61.2 & 41.8 
    &- &- &- &- &- &- \\ 
WaveMLP-S \cite{tang2021imagewavemlp} & 47 & 250
    & 44.0 & 65.8 & 48.2 & 40.0 & 63.1 & 42.9 
    &- &- &- &- &- &- \\ 
\oursxt & 35 & 215
    & 42.8 & 64.9 & 46.9 & 39.5 & 62.1 & 42.5
    & 45.0 & 67.4 & 49.5 & 41.1 & 64.4 & 44.2 \\ 
\arrayrulecolor{white}\hline\arrayrulecolor{black}
\ourst & 47 & 251
    & \textbf{44.8} & \textbf{66.9} & \textbf{49.0} & \textbf{41.0} & \textbf{64.2} & \textbf{44.3}
    & \textbf{47.1} & 69.0 & \textbf{51.7} & \textbf{42.7} & \textbf{66.5} & \textbf{46.0} \\ 

\midrule

Twins-L \cite{chu2021twins} & 120 & 474
    & 45.9 & - & - & 41.6 & - & -
    & - &- &- &- &- &-  \\
Swin-B \cite{liu2021swin} & 107 & 496
    & 45.5 & - & - & 41.3 & - & - 
    & 48.5 & 69.8 & 53.2 & 43.4 & 66.8 & 46.9  \\
MViT-B \cite{yan2022MViT} & 73 & 438 
    &- &- &- &- &- &-
    & 48.8 & 71.2 & 53.5 & 44.2 & 68.4 & 47.6 \\
CycleMLP-B5 \cite{chen2022cyclemlp} & 95 & 421 
    & 44.1 & 65.5 & 48.4 & 40.1 & 62.8 & 43.0
    &- &- &- &- &- &-   \\
WaveMLP-B \cite{tang2021imagewavemlp} & 75 & 353
    & 45.7  & 67.5 & 50.1 & 27.8 & 49.2 & 59.7
    &- &- &- &- &- &-  \\

\arrayrulecolor{white}
\hline
\arrayrulecolor{black}
\ourss & 58 & 311
    & 46.0 & 68.2 & 50.4 & 42.0 & 65.3 & 45.5
    & 48.1 & 69.7 & 52.9 & 43.4 & 67.2 & 46.7 \\
\oursb & 72 & 377
    & 46.5 & 68.6 & 51.0 & 42.5 & 66.1 & 45.8 
    & 49.0 & 70.7 & 54.0 & 43.9 & 67.7 & 47.5 \\
\arrayrulecolor{white}\hline\arrayrulecolor{black}
\oursl & 96 & 424
    & \textbf{47.4} & \textbf{69.9} & \textbf{52.0} & \textbf{43.2} & \textbf{67.3} & \textbf{46.5}
    & \textbf{49.5} & \textbf{71.5} & \textbf{54.3} & \textbf{44.5} & \textbf{68.7} & \textbf{48.1} \\

\bottomrule
\end{NiceTabular}
\captionof{table}{Object detection results on COCO \texttt{val2017} with Mask R-CNN \cite{he2017maskrcnn} 1$\times$ and 3$\times$ MS. FLOPS are evaluated with resolution 800$\times$1280. }
\label{table: coco_mask_all}
\end{center}
\end{table*}
\begin{table*}[h]
\begin{center}
\small
\begin{NiceTabular}{@{}lcccccccccccccc@{}}[code-before=
\rectanglecolor{\rc}{8-1}{9-15}
\rectanglecolor{\rc}{13-1}{15-15}
]
\toprule
\multirow{2}{*}{Backbone} & \Block{2-1}{\makecell{\#Params.\\(M)}} & \multirow{2}{*}{\makecell{FLOPs\\{(G)}}} & \multicolumn{6}{c}{\normalsize Cascade Mask R-CNN 1$\times$} & \multicolumn{6}{c}{\normalsize Cascade Mask R-CNN 3$\times$ MS} \\
& & & \textbf{AP$^b$} & AP$^b_{50}$ & AP$^b_{75}$ & \textbf{AP$^m$} & AP$^m_{50}$ & AP$^m_{75}$ & \textbf{AP$^b$} & AP$^b_{50}$ & AP$^b_{75}$ & \textbf{AP$^m$} & AP$^m_{50}$ & AP$^m_{75}$ \\

\midrule

Deit-S \cite{touvron2021deit} & 80 & 889 
    &- &- &- &- &- &-
    & 48.0 & 67.2 & 51.7 & 41.4 & 64.2 & 44.3 \\ 
Swin-T \cite{liu2021swin} & 86 & 745
    & 48.1 & 67.1 & 52.2 & 41.7 & 64.4 & 45.0
    & 50.4 & 69.2 & 54.7 & 43.7 & 66.6 & 47.3 \\ 
ConvNeXt-T \cite{liu2022convnext} & 86 & 741 
    &- &- &- &- &- &-
    & 50.4 & 69.1 & 54.8 & 43.7 & 66.5 & 47.3 \\
AS-MLP-T \cite{lian2021asmlp} & 86 & 739
    &- &- &- &- &- &-
    & 50.1 & 68.8 & 54.3 & 43.5 & 66.3 & 46.9 \\ 
Shift-T \cite{wang2022shift} & 86 & 743
    &- &- &- &- &- &-
    & 50.3 & 68.6 & 54.7 & 43.4 & 66.2 & 47.0 \\
\oursxt & 73 & 693
    & 47.6 & 66.5 & 51.4 & 41.3 & 63.6 & 44.6
    & 50.0 & 68.7 & 54.4 & 43.5 & 66.2 & 46.8 \\
\ourst & 85 & 730
    & \textbf{49.4} & \textbf{68.3} & \textbf{53.4} & \textbf{42.8} & \textbf{65.7} & \textbf{45.9}
    & \textbf{51.3} & \textbf{70.4} & \textbf{55.6} & \textbf{44.6} & \textbf{67.7} & \textbf{48.4} \\ 
\midrule

Swin-B \cite{liu2021swin} & 145 & 982 
    & 50.8 & 70.1 & 55.1 & 43.9 & 67.4 & 47.4
    & 51.9 & 70.5 & 56.4 & 45.0 & 68.1 & 48.9 \\
ConvNeXt-B \cite{liu2022convnext} & 146 & 964
    &- &- &- &- &- &-
    & 52.7 & 71.3 & 57.2 & 45.6 & 68.9 & 49.5 \\
ASMLP-B \cite{lian2021asmlp} & 145 & 961 
    &- &- &- &- &- &-
    & 51.5 & 69.8 & 55.8 & 44.6 & 67.6 & 48.2 \\
\ourss & 96 & 790
    & 51.0 & 70.3 & 55.4 & 44.2 & 67.5 & 47.6
    & 51.5 & 70.3 & 55.9 & 44.8 & 68.0 & 48.6 \\
\arrayrulecolor{white}\hline\arrayrulecolor{black}
\oursb & 110 & 856 
    & 51.3 & 70.4 & 55.7 & 44.6 & 68.0 & 48.2 
    & 52.5 & 71.6 & 57.0 & 45.6 & 69.1 & 49.5 \\
\arrayrulecolor{white}\hline\arrayrulecolor{black}
\oursl & 134 & 902 
    & \textbf{51.8} & \textbf{70.9} & \textbf{56.3} & \textbf{44.8} & \textbf{68.4} & \textbf{48.4}
    & \textbf{53.3} & \textbf{72.1} & \textbf{57.9} & \textbf{46.3} & \textbf{69.8} & \textbf{50.4} \\

\bottomrule 
\end{NiceTabular}
\captionof{table}{Object detection results on COCO \texttt{val2017} with Cascade Mask R-CNN \cite{cai2018cascade} 1$\times$ and 3$\times$ MS. FLOPS are evaluated with resolution 800$\times$1280. 
}
\label{table: coco_cascade_all}
\end{center}
\end{table*}

\noindent
\subsubsection{Results}
We report the top-1 accuracy comparison between our \ours~with recent CNN-, Transformer- and MLP-based backbones in Table \ref{table: int_sota}, where all methods are categorized into different groups w.r.t. the model size (\#Parameters) and computation complexity (FLOPs). All our different variants achieve higher accuracy compared with the scale-comparable methods.  
1) Our \ours-T, -B, and -L variants outperform the prominent Transformer Swin-T, -S, and -B by +0.8\%, +0.3\% and +0.3\% with comparable parameters and FLOPs. For larger models, \oursl$\uparrow$~surpasses Swin-B$\uparrow$ with \textbf{-23\%} computation cost.
2) Our \ours~also surpasses all recent MLP-like backbones (ASMLP, CycleMLP, ViP, and \etcno). 
Compared with the recent CycleMLP mixing tokens in a \textit{deterministic and local} manner, our five variants outperforms the corresponding CycleMLP variants by +0.8\%, 0.5\%, +0.4\%, +0.5\% and +0.7\% respectively, with comparable computation cost.

Note that some MLP-like backbones (\egno, MLP-Mixer, ResMLP, gMLP, ResMLP, ViP, sMLP and \etcno) in Table \ref{table: int_sota} are not validated in downstream dense prediction tasks, where the most architectures are not compatible with various input resolutions. In contrast, our \ours~is capable of dealing with different input scales, and shows pronounced performance on dense prediction tasks, which will be shown in the following sections.

\subsection{Semantic Segmentation}
\label{sec: seg}

\noindent
\subsubsection{Settings}
The semantic segmentation is validated on the ADE20K \cite{zhou2019semanticADE20K}, which contains 20K training and 2K validation images. To compare with more existing backbone designs, we adopt the \ours~models initialized with the pretrained weights on \imntk~as the backbones for two widely used frameworks, Semantic FPN \cite{kirillov2019semanticFPN} and UperNet \cite{xiao2018unifiedupernet}.

For the experiments on Semantic FPN framework, we mainly follow the setting of PVT \cite{wang2021pyramidpvt} to train models for 40K iterations with the batchsize of 32 (8 GPUs with 4 images per GPU). We use AdamW \cite{loshchilov2017decoupledadamw} optimizer with the initial learning rate of $2e^{-4}$ and the weight decay of $0.01$. The image is randomly resized and cropped to 512$\times$512 for training. We set the stochastic depths as 0.1, 0.1, 0.2, 0.2 and 0.2 for \oursxt, -T, -S, -B and -L respectively.

For the experiments on UperNet framework, we mainly follow the setting of Swin \cite{liu2021swin} to train models for 160K steps with the batchsize of 16 (8 GPUs with 2 images per GPU). We adopt the AdamW \cite{loshchilov2017decoupledadamw} optimizer with the initial learning rate of $6e^{-5}$ and the weight decay of $0.01$. All models are trained with the input size of 512$\times$512. We set the stochastic depths as 0.1, 0.1, 0.2, 0.2 and 0.2 for \oursxt, -T, -S, -B and -L respectively. We adopt \miou~and Multi-Scale ([0.5, 0.75, 1.0, 1.25, 1.5, 1.75]) \miou~as the evaluation metrics.

\noindent
\subsubsection{Results}
The results on top of UperNet and Semantic FPN are shown in Table \ref{table: ade_fpn} and Table \ref{table: ade_upernet}. For different model scales, \ours~outperforms all previous methods with comparable computation costs. The largest \oursl~with Sematic FPN outperforms previous \sota~Twins-L by \textbf{+1.4} \miou~with \textbf{-23\%} parameters and \textbf{-16\%} FLOPs.
\oursl~also achieves the new \sota~(\textbf{51.1} ms \miou) with UperNet, which surpasses the representative network Swin-B by \textbf{+1.4} \miou~with \textbf{-10\%} parameters. Note that \ourss~achieves comparable performance with Swin-B, but only requires about \textbf{-50\%} parameters.


It also shows that most previous MLP-like backbones (\egno, CycleMLP, ASMLP, MorphMLP) perform better than Transformer-based Swin/Twins for smaller models, but lag behind them for larger models. These \textit{manually designed} token mixing methods within them leads to remarkable limitations in exploring rich feature patterns, while the global-scope attention in Transformers allows extracting better features as model scaling up.
In contrast, \ours~shows its strong capability and scalability on segmentation over different model scales, especially for the large-scale models. The superiority of \ours~lies in the flexibility of \ourcore, which provides great capability to exploit sufficient features from visual signals with various scales and deformations, especially for the pixel-level tasks heavily relying on spatial information interaction.

\subsection{Object Detection}
\label{sec: det}


\noindent
\subsubsection{Settings}
The object detection is evaluated on the COCO \cite{lin2014microsoftcoco} dataset, which contains 118K and 5K images for the training and validation. For thorough comparison, we adopt our \imntk~pretrained \ours~as the backbone on top of three representative detection frameworks, \ieno, Mask R-CNN \cite{he2017maskrcnn}, RetinaNet \cite{lin2017FocallossRetinaNet} and Casecade Mask R-CNN \cite{cai2018cascade,he2017maskrcnn}.

Following the common practices in this field \cite{liu2021swin,dong2021cswin,chen2022cyclemlp,wang2021pyramidpvt,tang2021imagewavemlp}, we report the standard 1$\times$ (MS) schedule and 3$\times$ MS schedule detection results on COCO 2017 \texttt{val} for different frameworks. For 1$\times$ schedule, we train the model with the single-scale inputs for 12 epochs with the learning rate decayed at 8 and 11 epochs. The image is resized to the shorter side of 800 pixels, while the longer side does not exceed 1333 pixels. For 3$\times$ schedule, we train the model for 36 epochs with the learning rate decayed at 27 and 33 epochs. For MS training, the image is resized to the shorter side between 480 and 800 while the longer side no longer than 1333. All models are trained using AdamW \cite{loshchilov2017decoupledadamw} optimizer with the batch size of 16 (8 GPUs with 2 images per GPU). Stochastic depths are set as 0.1, 0.1, 0.3, 0.5 and 0.6 for \oursxt, -T, -S, -B and -L, respectively, to avoid overfitting.


\noindent
\subsubsection{Results} 
\noindent
The object detection results for Mask R-CNN 1$\times$ and 3$\times$(MS)
are shown in Table \ref{table: coco_mask_all}. 
Thanks to \ourcore's flexibility and effectiveness for token mixing, our \ours~obtains promising results on the challenging object detection. \ours~achieves the \sota~for the most model scales with different detectors.
For the Mask R-CNN 1$\times$ setting, our different model variants outperform the corresponding parameter-comparable Swin variants by \textbf{+2.6/+1.9} and \textbf{+1.9/+1.9} mAP$^b$/mAP$^m$ respectively, which demonstrates the \ours's superiority on dense prediction task, where the input is usually with larger resolution. 
For the largest models, \oursl~surpasses the \sota~ Twins-L by \textbf{+1.5} mAP$^b$ with \textbf{-20\%} parameters. The comparisons with RetinaNet 1$\times$/3$\times$ and Cascade Mask R-CNN 1$\times$/3$\times$ can be found in Table \ref{table: coco_retina_all} and Table \ref{table: coco_cascade_all}.

\begin{table}[h]
\footnotesize
\begin{center}
\begingroup

\setlength{\tabcolsep}{4pt}
\begin{NiceTabular}{@{}l@{\hskip3pt}cclcl@{}}[code-before=
\rectanglecolor{\rc}{4-2}{4-6}
\rectanglecolor{\rc}{6-2}{6-6}
\rectanglecolor{\rc}{8-2}{8-6}
\rectanglecolor{\rc}{10-2}{10-6}
]

\toprule
& & \multicolumn{2}{c}{Semantic FPN} & \multicolumn{2}{c}{Mask R-CNN} \\
Backbone & Neck & FLOPS & \multicolumn{1}{c}{mIoU} & FLOPS & \multicolumn{1}{c}{AP$^b$} \\

\midrule

\multirow{2}{*}{ResNet-50}
& FPN         & 45.9 & 37.3
            & 259.8 & 38.0 \\
& ATMFPN  
    & 48.9 & 40.3\tcg{\large$_{\uparrow3.0}$} 
    & 298.9 & 39.9\tcg{\large$_{\uparrow1.9}$}   \\

\arrayrulecolor{white}
\midrule
\arrayrulecolor{black}

\multirow{2}{*}{Swin-Tiny}
& FPN       & 47.5 & 41.5
            & 267.0 & 42.2 \\
& ATMFPN  
    & 47.5 & 43.7\tcg{\large$_{\uparrow2.2}$}
    & 267.1 & 43.5\tcg{\large$_{\uparrow1.3}$}   \\

\arrayrulecolor{white}
\midrule
\arrayrulecolor{black}

\multirow{2}{*}{\ourst}
& FPN         & 42.4 & 45.8
            & 251.1 & 44.8 \\
& ATMFPN  
    & 41.4 & 46.5\tcg{\large$_{\uparrow0.7}$}
    & 247.0 & 45.6\tcg{\large$_{\uparrow0.8}$}   \\

\arrayrulecolor{white}
\midrule
\arrayrulecolor{black}

\multirow{2}{*}{\oursl}
& FPN   
        & 86.6 & 48.1
        & 423.7 & 47.4 \\
& ATMFPN  
        & 86.6 & 48.3\tcg{\large$_{\uparrow0.2}$}
        & 423.8 & 48.4\tcg{\large$_{\uparrow1.0}$}   \\
\bottomrule

\end{NiceTabular}
\endgroup
\captionof{table}{FPN/ATMFPN for semantic segmentation with Semantic FPN and object detection with Mask R-CNN 1$\times$.
}
\label{table: atm_fpn}
\end{center}
\end{table}



\subsection{ATMFPN}
Our proposed \ourcore~can be adopted not only for constructing vision backbones, but also as an enhanced alternative for convolution-based decoders.
Based on FPN \cite{lin2017FPN}, we build an ATMFPN neck with ATM, and report the results on different backbones for object detection and semantic segmentation in Table \ref{table: atm_fpn}. 
With comparable computation cost, ResNet-50 with ATMFPN outperforms the na\"ive FPN by \textbf{+3.0} mIoU/\textbf{+1.9} AP$^b$ for segmentation and object detection respectively. ATMFPN also helps improve the performance for the backbone of Swin and \ours.
Thanks to the flexibility, our proposed \ourcore~is basically applicable for extracting better visual feature representations.

\subsection{Ablation Study and Analysis}
\label{sec: abla}

\noindent
\subsubsection{Effectiveness of ATM} Table~\ref{table: ablation} shows our ablation results.
In the baseline \ding{172} of \ours, all offsets are fixed to $0$, which means there is no spatial information interaction between different tokens in \ding{172}. This baseline achieves 79.3\%  accuracy on ImageNet-1K while its performance on dense prediction tasks is severely bounded due to the lack of adequate spatial interaction.
This also validates that token mixing is sorely vital for dense prediction tasks. With our proposed ATMNet w/o PEG (\ding{173}), the classification accuracy is improved by \textbf{+2.7\%}, and the performance on the dense prediction task is \textit{significantly improved by a large margin} (+\textbf{7.4} mAP$^b$ on COCO and \textbf{+7.7} mIoU on ADE20K). Our proposed \ourcore~brings sufficient information mixing to help extract more powerful features with negligible additional computation overhead. The PEG module is introduced for providing position information for offset generation, which 
helps a little for dense tasks.


\begin{table}[t]
\footnotesize
\begin{center}
\sethlcolor{lblue}



\setlength{\tabcolsep}{4.5pt}
\begin{NiceTabular}{@{}c@{\hspace{5pt}}cclll@{}}[code-before=
\rectanglecolor{\rc}{4-1}{4-6}
]
\toprule
ID & Model & FLOPs & INT & COCO & ADE20K \\

\ding{172} & Baseline & 3.890 & 79.3  & 36.0  & 37.9 \\
\ding{173} & ATMNet w/o PEG & 3.924 &
    82.0\color{ForestGreen}{\large$_{\uparrow2.7}$} &
    43.4\color{ForestGreen!85}{\large$_{\uparrow7.4}$} &
    45.6\color{ForestGreen!85}{\large$_{\uparrow7.7}$} \\
\ding{174} & ATMNet & 3.972 &
    82.0\tcg{\large$_{\uparrow2.7}$} &
    43.6\tcg{\large$_{\uparrow7.6}$} &
    45.8\tcg{\large$_{\uparrow7.9}$} \\
\bottomrule
\end{NiceTabular}

\captionof{table}{Ablation study. FLOPs are obtained on \imntk~with input resolution of 224$\times224$. COCO: AP$^b$ for RetinaNet 1$\times$. ADE20K: mIoU for Semantic FPN.}
\label{table: ablation}
\end{center}
\end{table}

\begin{figure}[t]
\centering
\includegraphics[width=0.48\textwidth]{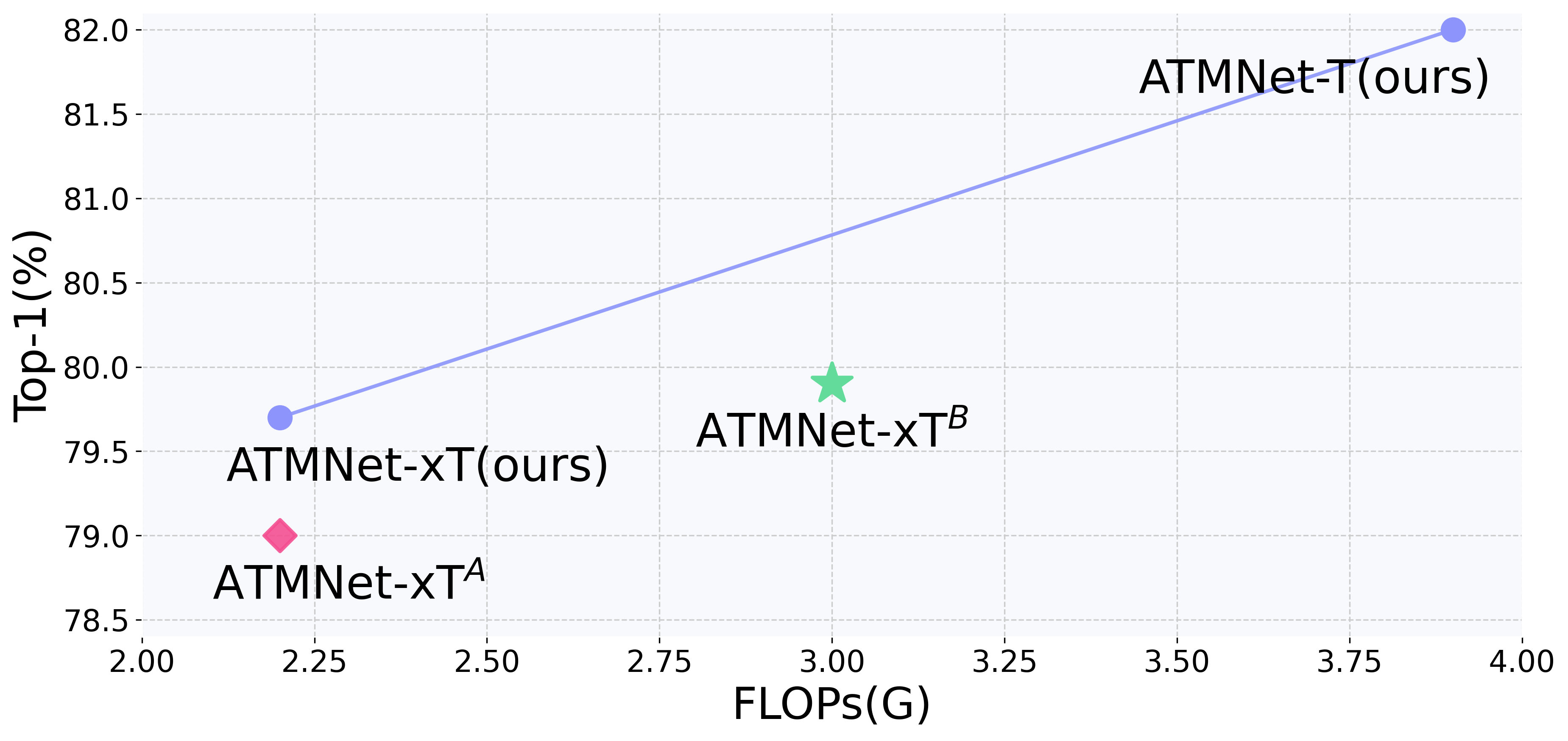}
\caption{Comparisons of different offset configurations on \imntk. 
\ours-xT$^A$(\textcolor{magenta}{$\blacklozenge$}): offset learning is not decoupled along different directions, \ieno, the selected contextual tokens are directly recomposed as $\mathbf{\tilde{x}}\!=\![  \mathbf{X}_{\left[ i + oh_1, j + ow_1, 1 \right]}, \mathbf{X}_{\left[ i + oh_2, j + ow_2, 2 \right]}, \dots ]^T$ with the predicted offsets $\{oh_c, ow_c\}$. 
\ours-xT$^B$(\textcolor{ForestGreen}{$\bigstar$}): the number of selected contextual tokens per channel is extended from 1 to 3 for each direction.
}	
\label{fig: abalation}
\end{figure}

\begin{figure}[t]
\centering
\includegraphics[width=\linewidth]{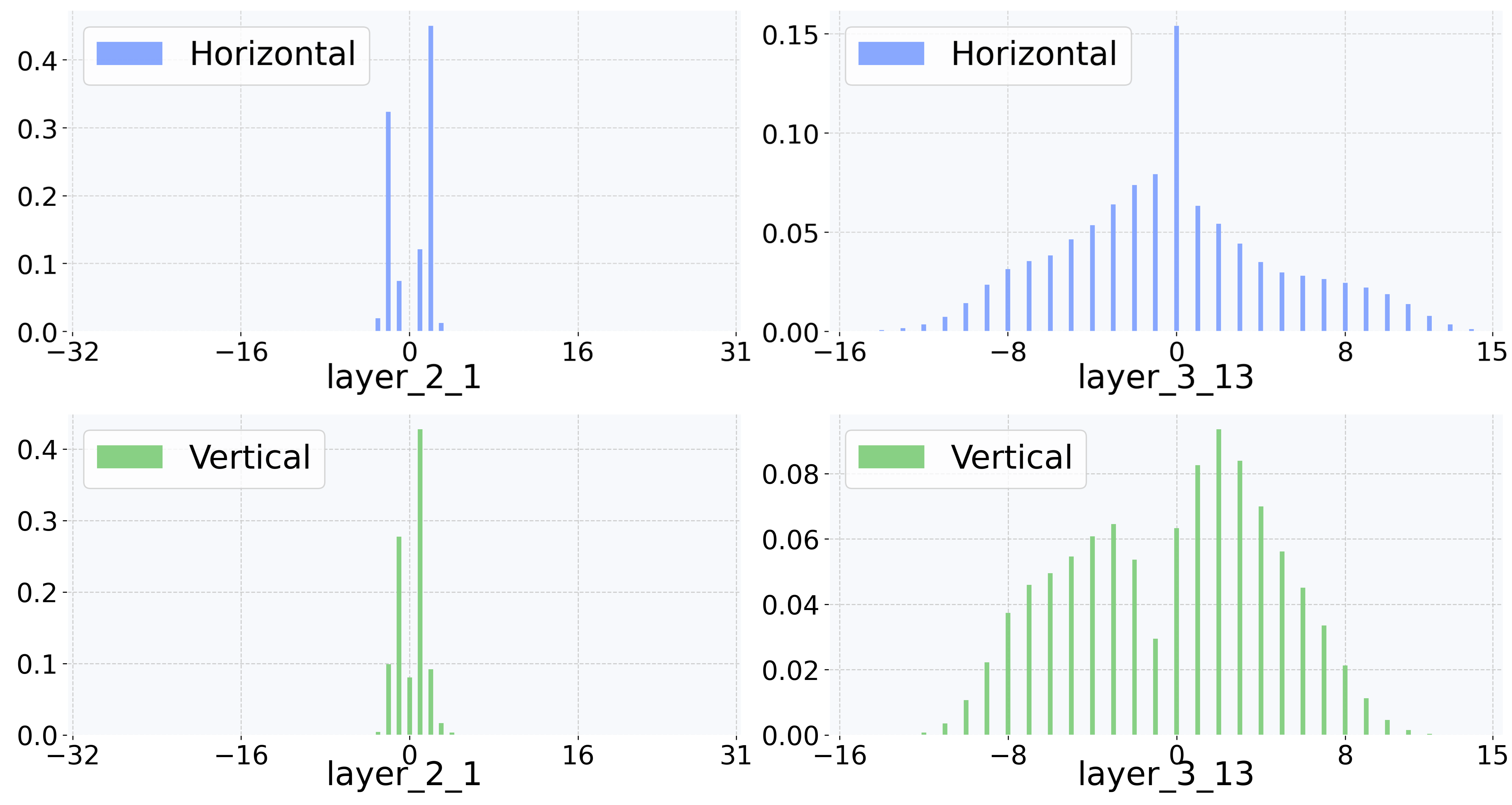}
\caption{
Histograms of learned offsets for the center token from different layers, counted on all samples from ADE20K~\texttt{val}. The range of horizontal axis is the possible maximal range for the corresponding feature resolution, \egno, $[-32, 31]$ for the first sub-figure. Input size: 512$\times$512. Feature resolutions for two columns: 64$\times$64 and 32$\times$32. \texttt{layer\_i\_j}: j$^{th}$ layer of i$^{th}$ stage.
}
\label{fig: offset_ade}
\end{figure}


\subsubsection{Comparison with other offset configurations} 1) \textit{Effectiveness of directional decomposition.} As show in Fig. \ref{fig: abalation}, our \ours-xT is clearly superior to \ours-xT$^A$(\textcolor{magenta}{$\blacklozenge$}) with very close FLOPs, demonstrating the effectiveness of directional decomposition during predicting offsets. 2) \textit{The number of selected contextual tokens.} The \ours-xT$^B$(\textcolor{ForestGreen}{$\bigstar$}) with more contextual tokens for each query outperforms \oursxt~by 0.2\% but with \textbf{+50\%} additional computation cost. This shows our \ourop~is a better trade-off between the computation cost and the final performance as an efficient and effective token mixer.

\subsubsection{Analyses of learned offsets}
We investigate the distributions of the learned offsets via the histograms of offsets w.r.t. the center token in Fig. \ref{fig: offset_ade}.
We observe: 1) As the depth increases, the learned offsets expand to a larger range. 
This is in line with the conclusion in \cite{raghu2021DoVisionTrans,zhang2021morphmlp} that local receptive fields in shallower layers are conductive to training vision models, while the long-range information is required for deeper layers. 
2) For a query token, the learned offsets differ for different channels and such flexibility enables efficient semantic-adaptive information interaction.
3) Besides the network depth, the learned offsets of ATM are also adaptive to different datasets or tasks (shown in the Supplementary), endowing \ours~with higher flexibility and better adaptivity. This observation indicates that mixing tokens with \textit{hand-crafted and deterministic} rules is in fact insufficient to model the various distributions of different datasets. 
More results are in the Supplementary.

\section{Conclusion}
In this work, we propose an innovative token mixing mechanism, \ourcore, which actively and meticulously learns to fuse content-adaptive contextual information in the global scope. With the proposed basic operator, we build a general vision backbone \ours~for various vision tasks and an enhanced FPN, \ieno, ATMFPN for dense prediction tasks. \ours~is capable of flexibly and effciently capturing diverse visual patterns. Comprehensive experiments demonstrate our \ours~is generally applicable and effective for various vision tasks including image classification, object detection and semantic segmentation. 
In future work, we will exploit \ours's potential of dealing with temporal video signals.


\clearpage
{
\small
\bibliographystyle{abbrv}
\bibliography{main}
}

\clearpage

\begin{figure*}[t]
	\centering
    \includegraphics[width=\linewidth]{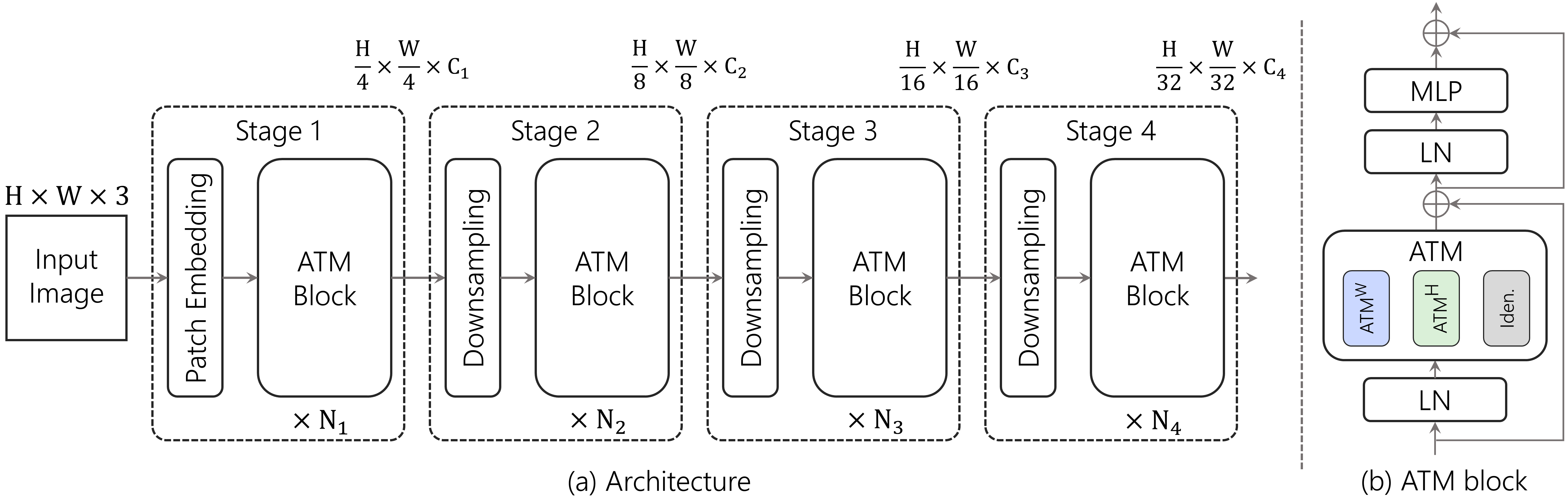}
    \caption{
	a) The overall architecture of \ours. b) The ATM block.
	}
	\label{fig: supp_arch}
\end{figure*}

\section*{A1\quad More Architecture Details}

Our \ours~consists of four stages with downsampling layers inserted between every two stages.The downsampling layer is implemented by overlapping patch embedding \cite{wang2021pyramidpvt,wang2021pvtv2} with stride of $2$, which reduces the spatial resolution in half. We illustrate the overall architecture of our \ours~in Fig. \ref{fig: supp_arch} (a), and show the basic ATM block in Fig. \ref{fig: supp_arch} (b). Each ATM block consists of LayerNorm (LN) \cite{ba2016LayerNorm}, ATM module, LN, and MLP sequentially with two skip connections. The MLP is composed of two Fully Connected (FC) layers which first increases the number of channels with a ratio $E_i$ then reduces it to the original one. 
We find that it is not necessary to generate new offsets for each ATM block, instead we generate new offsets every $K$ layers in each stage, where $K=2/2/6/6/6$ for the five variants \oursxt/T/S/B/L, respectively. 
The detailed configurations of different architecture variants are listed in Table \ref{table: supp_arch}.

\begin{table}[h]
\footnotesize
\begin{center}
\begingroup

\begin{NiceTabular}{c|c}
\hline
\Block{2-1}{Config} & Value \\
\cline{2-}
& xT, T, S, B, L \\
\hline

optimizer   & AdamW \\
learning rate & 1e-3 \\
weight decay & 0.05 \\ 
batch size  & 1024 \\
learning rate schedule & cosine decay \\
warmup epochs & 5 \\
training epochs & 300 \\
augmentation & \texttt{RandAug} (9, 0.5) \\
label smoothing \cite{szegedy2016InceptionV3} & 0.1 \\
mixup \cite{zhang2017mixup} & 0.8 \\
cutmix \cite{yun2019cutmix} & 1.0 \\
random erasing \cite{zhong2020randomerasing} & 0.25 \\
drop path \cite{huang2016deepdpr} & 0.1, 0.1, 0.2, 0.3, 0.3 \\
\hline

\end{NiceTabular}

\endgroup
\captionof{table}{
\imntk~training settings.
}
\label{table: supp_int_setting}
\end{center}
\end{table}

\begin{table*}[h]
\footnotesize
\begin{center}
\begingroup

\begin{NiceTabular}{@{}c|c|c|c|c|c|c|c@{}}

\hline
& Output size 
& Layer Name
& -xT & -T & -S & -B & -L \\
\hline

\multirow{5}{*}{Stage 1} 
& \Block{5-1}{$\frac{H}{4}\times\frac{W}{4}$}
& \multirow{2}{*}{\makecell{Overlapping\\Patch Embedding}}
& \Block{1-5}{$S_1=4$} & & & & \\
\cline{4-8}
& & & \Block{1-4}{$C_1=64$} & & & & $C_1=96$ \\
\cline{3-8}
& & \multirow{3}{*}{\makecell{ATM Block}}
& $E_1=4$ & $E_1=4$ & $E_1=8$ & $E_1=8$ & $E_1=4$ \\
& & & $N_1=2$ & $N_1=2$ & $N_1=3$ & $N_1=3$ & $N_1=3$ \\
& & & $|\mathcal{O}_1|=32$ & $|\mathcal{O}|_1=32$ & $|\mathcal{O}|_1=32$ & $|\mathcal{O}|_1=32$ & $|\mathcal{O}|_1=48$ \\ 
\hline

\multirow{5}{*}{Stage 2} 
& \Block{5-1}{$\frac{H}{8}\times\frac{W}{8}$}
& \multirow{2}{*}{\makecell{Downsampling}}
& \Block{1-5}{$S_2=2$} & & & & \\
\cline{4-8}
& & & \Block{1-4}{$C_2=128$} & & & & $C_2=192$ \\
\cline{3-8}
& & \multirow{3}{*}{\makecell{ATM Block}}
& $E_2=4$ & $E_2=4$ & $E_2=8$ & $E_2=8$ & $E_2=4$ \\
& & & $N_2=2$ & $N_2=3$ & $N_2=4$ & $N_2=8$ & $N_2=4$ \\
& & & $|\mathcal{O}|_2=32$ & $|\mathcal{O}|_2=32$ & $|\mathcal{O}|_2=32$ & $|\mathcal{O}|_2=32$ & $|\mathcal{O}|_2=48$ \\ 
\hline

\multirow{5}{*}{Stage 3} 
& \Block{5-1}{$\frac{H}{16}\times\frac{W}{16}$}
& \multirow{2}{*}{\makecell{Downsampling}}
& \Block{1-5}{$S_3=2$} & & & & \\
\cline{4-8}
& & & \Block{1-4}{$C_3=320$} & & & & $C_3=384$ \\
\cline{3-8}
& & \multirow{3}{*}{\makecell{ATM Block}}
& $E_3=4$ & $E_3=4$ & $E_3=4$ & $E_3=4$ & $E_3=4$ \\
& & & $N_3=4$ & $N_3=10$ & $N_3=18$ & $N_3=27$ & $N_3=24$ \\
& & & $|\mathcal{O}|_3=80$ & $|\mathcal{O}|_3=80$ & $|\mathcal{O}|_3=80$ & $|\mathcal{O}|_3=80$ & $|\mathcal{O}|_3=96$ \\ 
\hline

\multirow{5}{*}{Stage 4} 
& \Block{5-1}{$\frac{H}{32}\times\frac{W}{32}$}
& \multirow{2}{*}{\makecell{Downsampling}}
& \Block{1-5}{$S_4=2$} & & & & \\
\cline{4-8}
& & & \Block{1-4}{$C_4=512$} & & & & $C_4=768$ \\
\cline{3-8}
& & \multirow{3}{*}{\makecell{ATM Block}}
& $E_4=4$ & $E_4=4$ & $E_4=8$ & $E_4=8$ & $E_4=4$ \\
& & & $N_4=2$ & $N_4=3$ & $N_4=4$ & $N_4=8$ & $N_4=4$ \\
& & & $|\mathcal{O}|_4=64$ & $|\mathcal{O}|_4=64$ & $|\mathcal{O}|_4=64$ & $|\mathcal{O}|_4=64$ & $|\mathcal{O}|_4=96$ \\ 
\hline


\end{NiceTabular}

\endgroup
\captionof{table}{
Architecture variants. $C_i$, $E_i$, $N_i$ and $|\mathcal{O}_i|$ denote the number of channels, channel expand ratio, number of ATM blocks, and number of offsets per ATM block for the $i^{th}$ stage. 
}
\label{table: supp_arch}
\end{center}
\end{table*}

\begin{figure*}[h]
	\centering
    \includegraphics[width=0.9\linewidth]{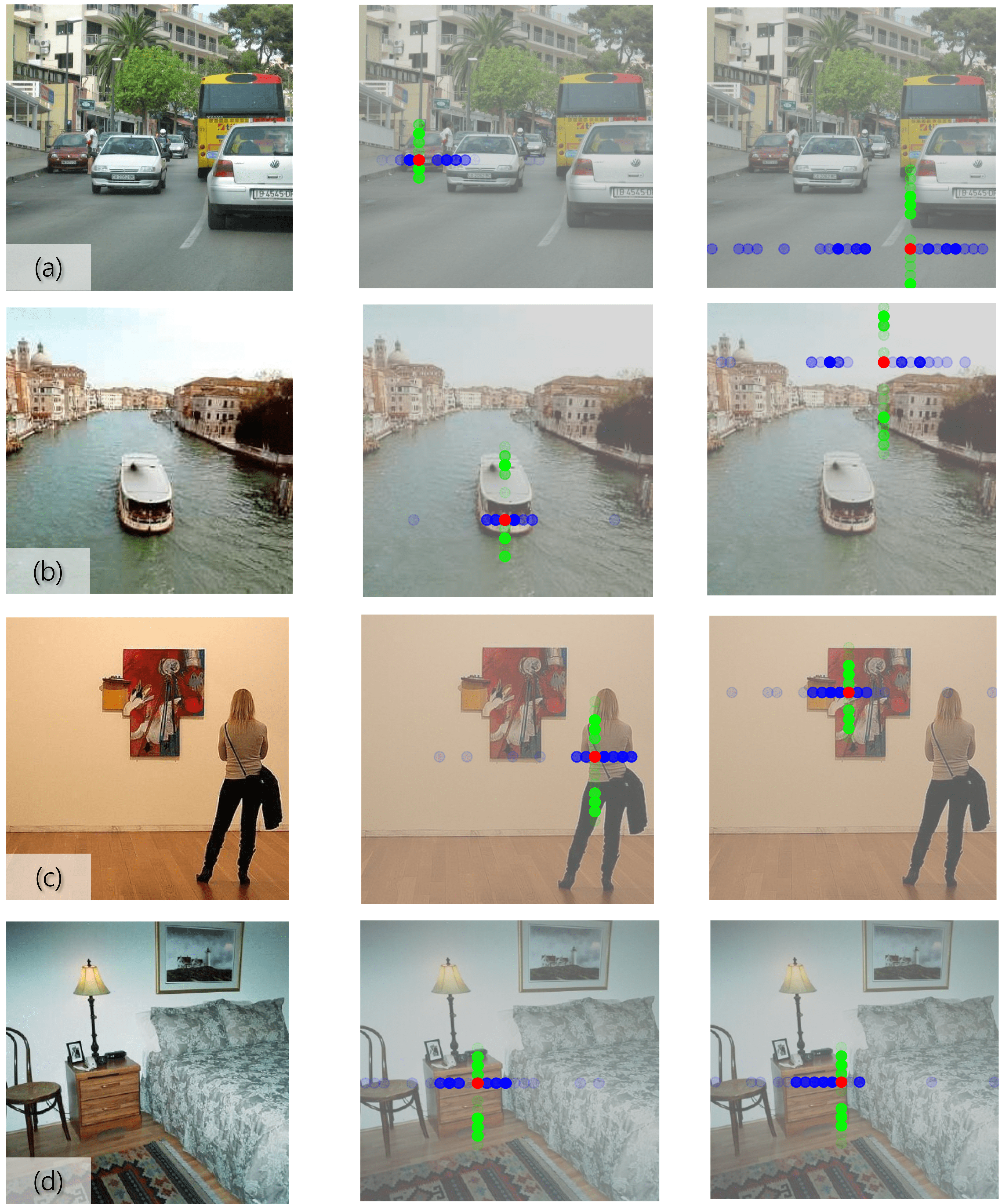}
    \caption{
	Illustration of the horizontal (\tikzcircle[blue!80, fill=blue!80]{3pt}) and vertical (\tikzcircle[green!80, fill=green!80]{3pt}) offsets for the given query token (\tikzcircle[red!80, fill=red!80]{3pt}) on ADE20K (input size: 512$\times$512). 
	The transparency of each circle corresponds to how many times the token at this position is sampled, \ieno, the more transparent the circle is, the corresponding offset values appear less in $\mathcal{O}=\{ o_i \}_{i=1}^C$.
    The visualized offsets are from a randomly sampled layer \texttt{layer\_3\_18}. Similar phenomenons can be observed at other layers.
	}	
	\label{fig: supp_offset_1}
\end{figure*}

\begin{figure*}[h]
	\centering
	\begin{tabular}{@{}c@{}}
    \includegraphics[width=\linewidth]{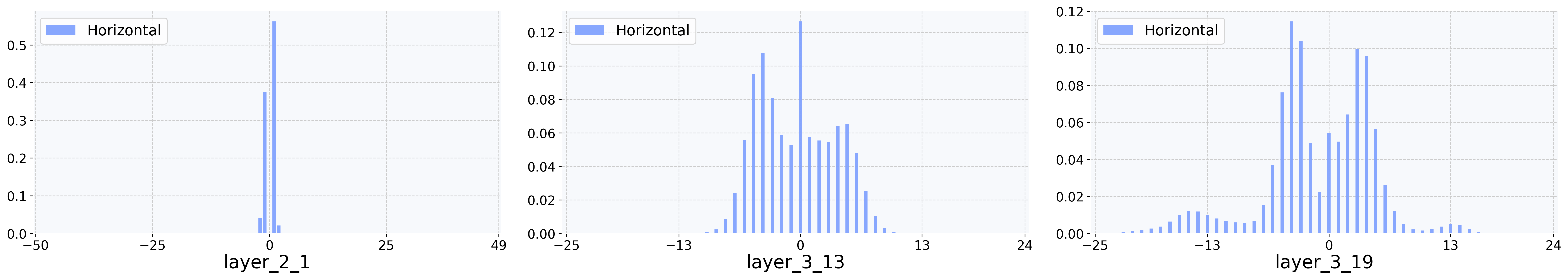}
    \end{tabular}
    \vspace{\floatsep}
    \begin{tabular}{@{}c@{}}
    \includegraphics[width=\linewidth]{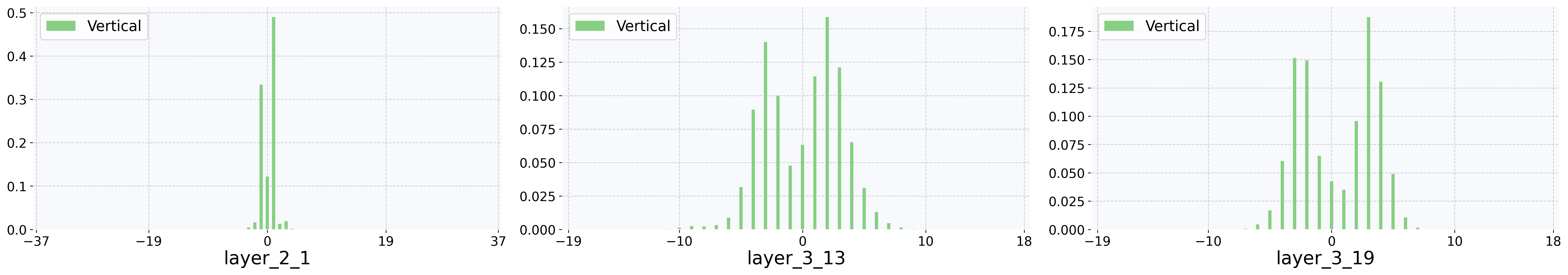}
    \end{tabular}
    \caption{
	Histograms of learned offsets for the center token from different layers, counted on all samples from COCO \texttt{val}. The range of horizontal axis is the possible maximal range for the corresponding feature resolution, \egno, $[-50, 49]$ for the first sub-figure. Input size: 600$\times$800. Feature resolutions for three columns: 75$\times$100, 38$\times$50 and 38$\times$50. \texttt{layer\_i\_j}: j$^{th}$ layer of i$^{th}$ stage.
	}	
	\label{fig: supp_offset_coco}
\end{figure*}

\begin{figure*}[h]
	\centering
	\begin{tabular}{@{}c@{}}
    \includegraphics[width=\linewidth]{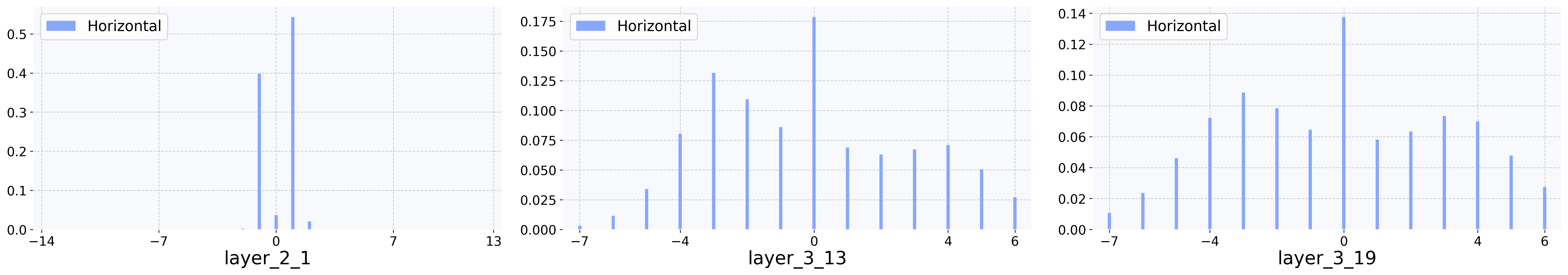}
    \end{tabular}
    \vspace{\floatsep}
    \begin{tabular}{@{}c@{}}
    \includegraphics[width=\linewidth]{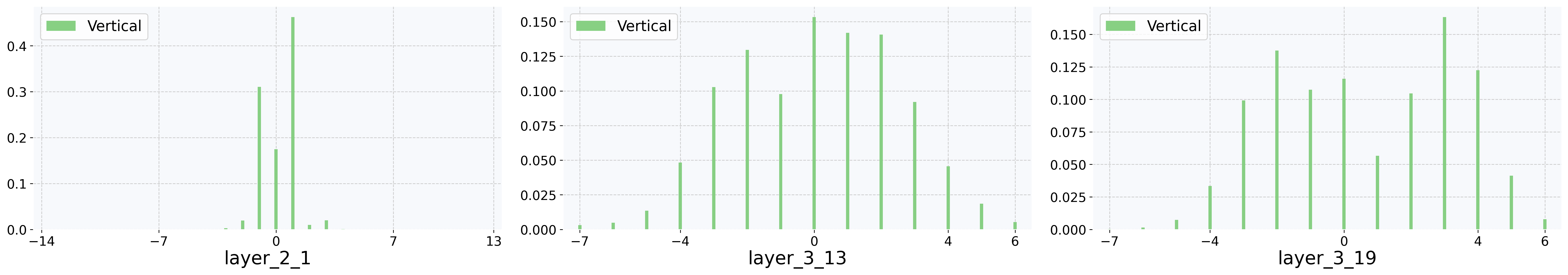}
    \end{tabular}
    \caption{
	Histograms of learned offsets for the center token from different layers, counted on 5K random samples from \imntk~\texttt{val}. The range of horizontal axis is the possible maximal range for the corresponding feature resolution, \egno, $[-14, 13]$ for the first sub-figure. Input size: 224$\times$224. Feature resolutions for three columns: 28$\times$28, 14$\times$14 and 14$\times$14. \texttt{layer\_i\_j}: j$^{th}$ layer of i$^{th}$ stage.
	}	
	\label{fig: supp_offset_int}
\end{figure*}

\section*{A2\quad More Experiment Results}

\subsection*{A2.1\quad Visualization of the Learned Offsets}
\label{sec: visu}

We visualize the learned offsets in \ours~for different input images in Fig. \ref{fig: supp_offset_1}. There are two main observations:
1) The learned offsets are adaptive to the \emph{scales} of visual contents. As illustrated in Fig. \ref{fig: supp_offset_1} (a) and (b), the token mixing is performed in a wider range for the query token located in large-scale object. 
2) The learned offsets are adaptive to the \emph{shapes} of visual contents. As illustrated in Fig. \ref{fig: supp_offset_1} (c) and (d), our \ours~tends to mix tokens within semantic-relevant regions, exhibiting obvious shape adaptability.

\subsection*{A2.2\quad Statistics of the Learned Offsets}
\label{sec: stat}
We investigate the distributions of the learned offsets at different depths, and show the histograms of offsets w.r.t. the center token on COCO (object detection task) in Table \ref{fig: supp_offset_coco}. The histograms are counted over all samples from COCO \texttt{val}. 
As the depth increases, the learned horizontal/vertical offsets spread to larger receptive field. This is in line with the conclusion in \cite{raghu2021DoVisionTrans,zhang2021morphmlp} that local receptive fields in shallower layers are conductive to training vision models, while the long-range information is required for deeper layers.

We also present the distributions of the learned offsets on \imntk~(image classification task) in Fig. \ref{fig: supp_offset_int}. Besides the adaptivity to the network depth, we observe that the learned offsets of ATM are also adaptive to different datasets or tasks, endowing \ours~with higher flexibility and better adaptivity. This observation also indicates that mixing tokens with hand-crafted and deterministic rules \cite{chen2022cyclemlp,zhang2021morphmlp,hou2021visionvip} is in fact insufficient to model the various distributions of different datasets.

\end{document}